\documentclass[11pt,a4paper]{article}
\usepackage{authblk}

\usepackage[hyperref]{emnlp2019}
\usepackage{times}
\usepackage{latexsym}

\usepackage{url}
\usepackage{graphicx}
\usepackage{amsmath}
\usepackage{amsfonts}
\usepackage{mathtools}
\usepackage{array}

\usepackage{caption}
\usepackage{subcaption}
\usepackage{fancyhdr}
\fancypagestyle{firststyle}
{
   \fancyhf{}
   \fancyfoot[C]{Proceedings of the 2019 Conference on Empirical Methods in Natural Language Processing and the 9th International Joint Conference on Natural Language Processing (EMNLP-IJCNLP 2019), Hong Kong, China.}
}

\aclfinalcopy 

\newcolumntype{A}{>{\raggedright\arraybackslash}m{6cm}}
\newcolumntype{B}{>{\raggedright\arraybackslash}m{8.5cm}}

\title{Improving Relation Extraction with Knowledge-attention}

\author[1]{Pengfei Li}
\author[1\Thanks{ Corresponding author.}]{Kezhi Mao}
\author[3]{Xuefeng Yang}
\author[2]{Qi Li}
\affil[1]{School of Electrical and Electronic Engineering, Nanyang Technological University, Singapore}
\affil[2]{Interdisciplinary Graduate School, Nanyang Technological University, Singapore}
\affil[3]{ZhuiYi Technology, Shenzhen, China}
\affil[ ]{\tt{\{pli006,ekzmao,liqi0024\}@ntu.edu.sg, ryan@wezhuiyi.com}}
 
\date{}

\begin{document}
\maketitle
\thispagestyle{firststyle}

\begin{abstract}
  While attention mechanisms have been proven to be effective in many NLP tasks, majority of them are data-driven. We propose a novel knowledge-attention encoder which incorporates prior knowledge from external lexical resources into deep neural networks for relation extraction task. Furthermore, we present three effective ways of integrating knowledge-attention with self-attention to maximize the utilization of both knowledge and data. The proposed relation extraction system is end-to-end and fully attention-based. Experiment results show that the proposed knowledge-attention mechanism has complementary strengths with self-attention, and our integrated models outperform existing CNN, RNN, and self-attention based models. State-of-the-art performance is achieved on TACRED, a complex and large-scale relation extraction dataset.
\end{abstract}

\section{Introduction}
Relation extraction aims to detect the semantic relationship between two entities in a sentence. 
For example, given the sentence: 
{\small``\emph{James Dobson} has resigned as chairman of \emph{Focus On The Family}, which he founded thirty years ago."}, the goal is to recognize the organization-founder relation held between {\small ``\emph{Focus On The Family}"} and {\small ``\emph{James Dobson}"}. 
The various relations between entities extracted from large-scale unstructured texts can be used for ontology and knowledge base population~\cite{chen2018on2vec,fossati2018n}, as well as
facilitating downstream tasks that requires relational understanding of texts such as question answering~\cite{yu2017improved} and dialogue systems~\cite{young2018augmenting}.

Traditional feature-based and kernel-based approaches require extensive feature engineering~\cite{suchanek2006combining, qian2008exploiting, rink2010utd}. Deep neural networks such as Convolutional Neural Networks (CNNs) and Recurrent Neural Networks (RNNs) have the ability of exploring more complex semantics and extracting features automatically from raw texts for relation extraction tasks~\cite{xu2016improved, vu2016combining, lee2017semeval}.
Recently, attention mechanisms have been introduced to deep neural networks to improve their performance~\cite{zhou2016attention, wang2016relation,zhang2017position}. Especially, the Transformer proposed by Vaswani et al.~\shortcite{vaswani2017attention} is based solely on self-attention and has demonstrated better performance than traditional RNNs~\cite{bilan2018position,verga2018simultaneously}. However, deep neural networks normally require sufficient labeled data to train their numerous model parameters. The scarcity or low quality of training data will limit the model's ability to recognize complex relations and also cause overfitting issue.

A recent study \cite{li2019knowledge} shows that incorporating prior knowledge from external lexical resources into deep neural network can reduce the reliance on training data and improve relation extraction performance. Motivated by this, we propose a novel knowledge-attention mechanism, which transforms texts from word semantic space into relational semantic space by attending to relation indicators that are useful in recognizing different relations. The relation indicators are automatically generated from lexical knowledge bases which represent keywords and cue phrases of different relation expressions.
While the existing self-attention encoder learns internal semantic features by attending to the input texts themselves, the proposed knowledge-attention encoder captures the linguistic clues of different relations based on external knowledge. Since the two attention mechanisms complement each other, we integrate them into a single model to maximize the utilization of both knowledge and data, and achieve optimal performance for relation extraction. 

In summary, the main contributions of the paper are: (1) We propose knowledge-attention encoder, a novel attention mechanism which incorporates prior knowledge from external lexical resources to effectively capture the informative linguistic clues for relation extraction. 
(2) To take the advantages of both knowledge-attention and self-attention, we propose three integration strategies: multi-channel attention, softmax interpolation, and knowledge-informed self-attention. Our final models are fully attention-based and can be easily set up for end-to-end training. 
(3) We present detailed analysis on knowledge-attention encoder. Results show that it has complementary strengths with self-attention encoder, and the integrated models achieve start-of-the-art results for relation extraction.

\section{Related Works}
We focus here on deep neural networks for relation extraction since they have demonstrated better performance than traditional feature-based and kernel-based approaches.

Convolutional Neural Networks (CNNs) and Recurrent Neural Networks (RNNs) are the earliest and commonly used approaches for relation extraction. Zeng et al.~\shortcite{zeng2014relation} showed that CNN with position embeddings is effective for relation extraction. Similarly, CNN with multiple filter sizes~\cite{nguyen2015relation}, pairwise ranking loss function~\cite{dos2015classifying} and auxiliary embeddings~\cite{lee2017semeval} were proposed to improve performance. Zhang and Wang~\shortcite{zhang2015relation} proposed bi-directional RNN with max pooling to model the sequential relations. Instead of modeling the whole sentence, performing RNN on sub-dependency trees (e.g. shortest dependency path between two entities) has demonstrated to be effective in capturing long-distance relation patterns~\cite{xu2016improved, miwa2016end}. Zhang et al.~\shortcite{zhang2018graph} proposed graph convolution over dependency trees and achieved state-of-the-art results on TACRED dataset.

Recently, attention mechanisms have been widely applied to CNNs~\cite{wang2016relation, han2018hierarchical} and RNNs~\cite{zhou2016attention, zhang2017position, du2018multi}. The improved performance demonstrated the effectiveness of attention mechanisms in deep neural networks.
Particularly, Vaswani et al.~\shortcite{vaswani2017attention} proposed a solely self-attention-based model called Transformer, which is more effective than RNNs in capturing long-distance features since it is able to draw global dependencies without regard to their distances in the sequences. Bilan and Roth~\shortcite{bilan2018position} first applied self-attention encoder to relation extraction task and achieved competitive results on TACRED dataset.  Verga et al.~\shortcite{verga2018simultaneously} used self-attention to encode long
contexts spanning multiple sentences for biological relation extraction. 
However, more attention heads and layers are required for self-attention encoder to capture complex semantic and syntactic information since learning is solely based on training data. Hence, more high quality training data and computational power are needed. 
Our work utilizes the knowledge from external lexical resources to improve deep neural network's ability of capturing informative linguistic clues.

External knowledge has shown to be effective in neural networks for many NLP tasks. Existing works focus on utilizing external knowledge to improve embedding representations~\cite{chen2015revisiting, liu2015learning, sinoara2019knowledge}, CNNs~\cite{toutanova2015representing, wang2017combining, li2019knowledge}, and RNNs~\cite{ahn2016neural, chen2016knowledge, chen2018neural, shen2018knowledge}. 
Our work is the first to incorporate knowledge into Transformer through a novel knowledge-attention mechanism to improve its performance on relation extraction task.

\section{Knowledge-attention Encoder}
We present the proposed knowledge-attention encoder in this section. Relation indicators are first generated from external lexical resources (Section \ref{RI_generation}); Then the input texts are transformed from word semantic space into relational semantic space by attending to the relation indicators using knowledge-attention mechanism (Section \ref{knowledge_attn}); Finally, position-aware attention is used to summarize the input sequence by taking both relation semantics and relative positions into consideration (Section \ref{position_attn}).

\begin{figure*}[t]
\centering
\begin{subfigure}{.5\textwidth}
  \centering
  \includegraphics[width=0.9\linewidth]{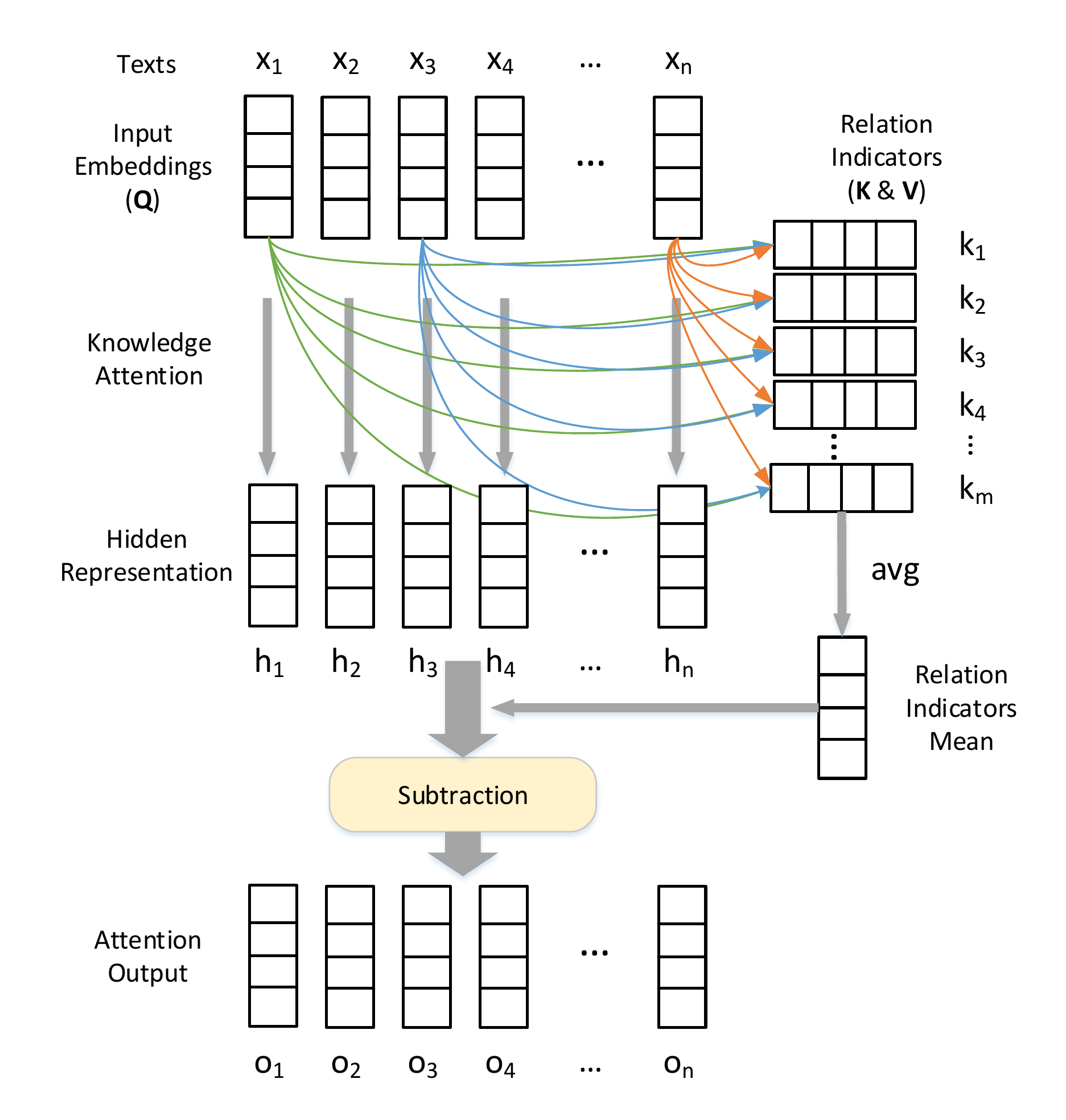}
  \label{fig1_sub1}
\end{subfigure}%
\begin{subfigure}{.4\textwidth}
  \centering
  \includegraphics[width=0.74\linewidth]{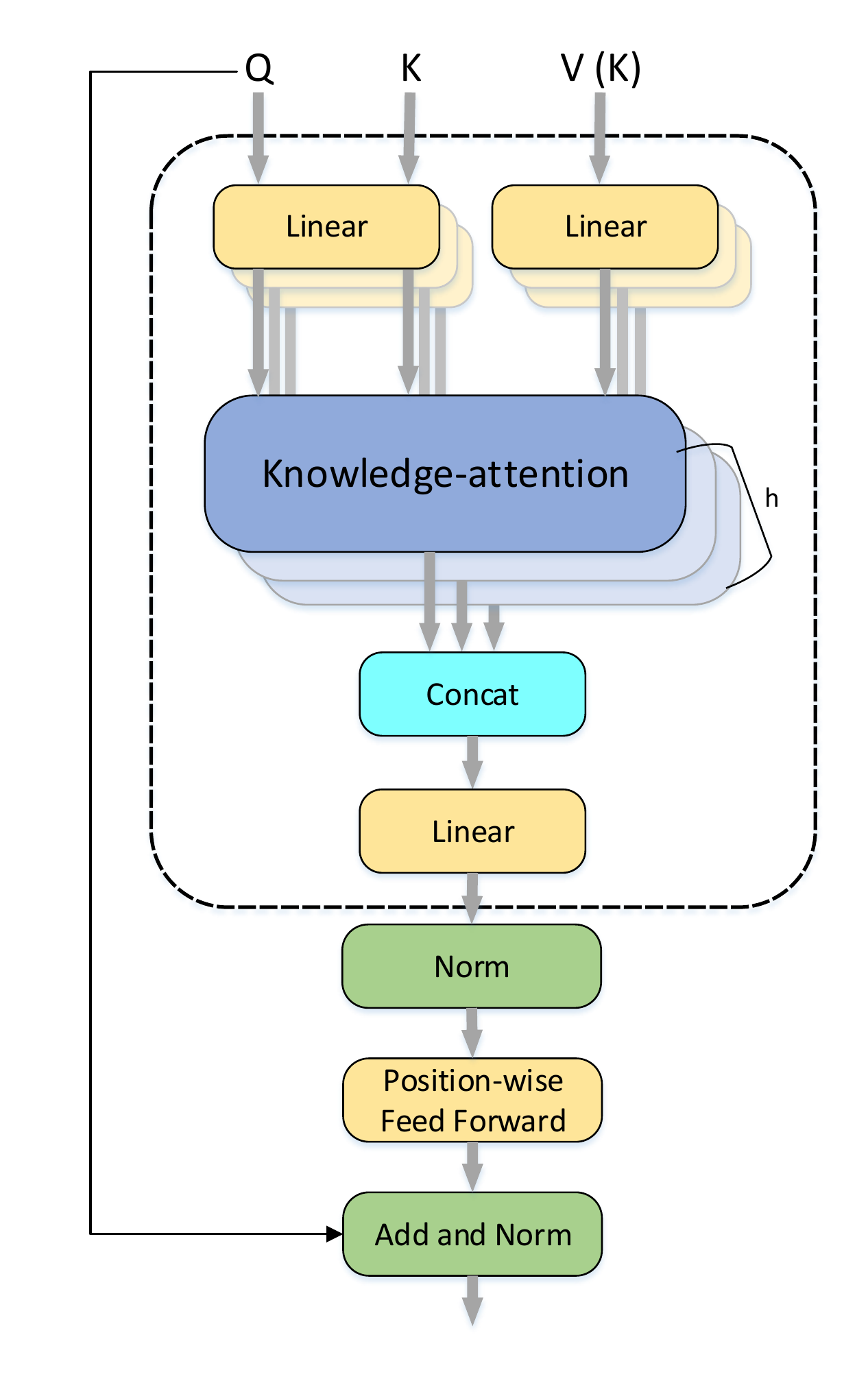}
  \label{fig1_sub2}
\end{subfigure}
\caption{Knowledge-attention process (left) and multi-head structure (right) of knowledge-attention encoder.}
\label{fig1_architecture}
\end{figure*}

\subsection{Relation Indicators Generation} \label{RI_generation}
Relation indicators represent the keywords or cue phrases of various relation types, which are essential for knowledge-attention encoder to capture the linguistic clues of certain relation from texts. 
We utilize two publicly available lexical resources including FrameNet\footnote{\url{https://framenet.icsi.berkeley.edu/fndrupal}} and Thesaurus.com\footnote{\url{https://www.thesaurus.com}} to find such lexical units.

FrameNet is a large lexical knowledge base which categorizes English words and sentences into higher level semantic frames~\cite{ruppenhofer2006framenet}. Each frame is a conceptual structure describing a type of event, object or relation.
FrameNet contains over 1200 semantic frames, many of which represent various semantic relations. 
For each relation type in our relation extraction task, we first find all the relevant semantic frames by searching from FrameNet (refer Appendix for detailed semantic frames used). Then we extract all the lexical units involved in these frames, which are exactly the keywords or phrases that often used to express such relation.
Thesaurus.com is the largest online thesaurus which has over 3 million synonyms and antonyms. It also has the flexibility to filter search results by relevance, POS tag, word length, and complexity. 
To broaden the coverage of relation indicators, we utilize the synonyms in Thesaurus.com to extend the lexical units extracted from FrameNet. To reduce noise, only the most relevant synonyms with the same POS tag are selected. 

Relation indicators are generated based on the word embeddings and POS tags of lexical units. Formally, given a word in a lexical unit, we find its word embedding $\mathbf{w}_i\in \mathbb{R}^{d_w}$ and POS embedding $\mathbf{t}_i\in \mathbb{R}^{d_t}$ by looking up the word embedding matrix $\mathbf{W}^{wrd}\in \mathbb{R}^{d_w\times V^{wrd}}$ and POS embedding matrix $\mathbf{W}^{pos}\in \mathbb{R}^{d_t\times V^{pos}}$ respectively, where $d_w$ and $d_t$ are the dimensions of word and POS embeddings, $V^{wrd}$ is vocabulary size\footnote{Same word embedding matrix is used for relation indicators and input texts, hence the vocabulary also includes all the words in the training corpus.} and $V^{pos}$ is total number of POS tags. The corresponding relation indicator is formed by concatenating word embedding and POS embedding, $\mathbf{k}_i=[\mathbf{w}_i, \mathbf{t}_i]$. If a lexical unit contains multiple words (i.e. phrase), the corresponding relation indicator is formed by averaging the embeddings of all words. Eventually, around 3000 relation indicators (including 2000 synonyms) are generated: $\mathbf{K}=\{\mathbf{k}_1,\mathbf{k}_2,...,\mathbf{k}_m \}$.

\subsection{Knowledge Attention} \label{knowledge_attn}

\subsubsection{Knowledge-attention process} \label{knowledge_attn1}
In a typical attention mechanism, a query ($q$) is compared with the keys ($K$) in a set of key-value pairs and the corresponding attention weights are calculated. The attention output is weighted sum of values ($V$) using the attention weights. In our proposed knowledge-attention encoder, the queries are input texts and the key-value pairs are both relation indicators. The detailed process of knowledge-attention is shown in Figure \ref{fig1_architecture} (left).

Formally, given text input $x=\{x_1,x_2,...,x_n\}$, the input embeddings  $\mathbf{Q}=\{\mathbf{q}_1,\mathbf{q}_2,...,\mathbf{q}_n \}$ are generated by  concatenating each word's word embedding and POS embedding in the same way as relation indicator generation in Section \ref{RI_generation}. The hidden representations $\mathbf{H}=\{\mathbf{h}_1,\mathbf{h}_2,...,\mathbf{h}_n \}$ are obtained by attending to the knowledge indicators $\mathbf{K}$, as shown in Equation \ref{attn_hidden}. The final knowledge-attention outputs are obtained by subtracting the hidden representations with the relation indicators mean, as shown in Equation \ref{attn_knwl}.
\begin{equation} \label{attn_hidden}
    \mathbf{H} = softmax(\frac{\mathbf{QK}^T}{\sqrt{d_k}})\mathbf{V}
\end{equation}
\begin{equation} \label{attn_knwl}
    knwl(\mathbf{Q},\mathbf{K},\mathbf{V}) = \mathbf{H}-\sum \mathbf{K}/m
\end{equation}
where $knwl$ indicates knowledge-attention process, $m$ is the number of relation indicators, and $d_k$ is dimension of key/query vectors which is a scaling factor same as in Vaswani et al.~\shortcite{vaswani2017attention}.

The subtraction of relation indicators mean will result in small outputs for irrelevant words. More importantly, the resulted output will be close to the related relation indicators and further apart from other relation indicators in relational semantic space. Therefore, the proposed knowledge-attention mechanism is effective in capturing the linguistic clues of relations represented by relation indicators in the relational semantic space.

\subsubsection{Multi-head knowledge-attention}
Inspired by the multi-head attention in Transformer~\cite{vaswani2017attention}, we also have multi-head knowledge-attention which first linearly transforms $\mathbf{Q}$, $\mathbf{K}$ and $\mathbf{V}$ $h$ times, and then perform $h$ knowledge-attentions simultaneously, as shown in Figure \ref{fig1_architecture} (right).

Different from the Transformer encoder, we use the same linear transformation for $\mathbf{Q}$ and $\mathbf{K}$ in each head to keep the correspondence between queries and keys. 
\begin{equation}
    head_i = knwl(\mathbf{Q}\mathbf{W}_i^Q, \mathbf{K}\mathbf{W}_i^Q, \mathbf{V}\mathbf{W}_i^V )
\end{equation}
where $\mathbf{W}_i^Q, \mathbf{W}_i^V \in \mathbb{R}^{d_k\times (d_k/h)}$ and $i \in [1,2,...h]$.
Besides, only one residual connection from input embeddings to outputs of position-wise feed forward networks is used. We also mask the outputs of padding tokens using zero vectors.

The multi-head structure in knowledge-attention allows the model to jointly attend inputs to different relational semantic subspaces with different contributions of relation indicators. This is beneficial in recognizing complex relations where various compositions of relation indicators are needed.

\subsection{Position-aware Attention} \label{position_attn}
It has been proven that the relative position information of each token with respective to the two target entities is beneficial for relation extraction task~\cite{zeng2014relation}.
We modify the position-aware attention originally proposed by Zhang et al.~\shortcite{zhang2017position} to incorporate such relative position information and find the importance of each token to the final sentence representation. 

Assume the relative position of token $x_i$ to target entity is $\hat{p_i}$. We apply position binning function (Equation \ref{binning_pos}) to make it easier for the model to distinguish long and short relative distances.

\begin{equation} \label{binning_pos}
    p_i=\left\{\begin{matrix}
            \hat{p_i} & \left | \hat{p_i} \right |\leq 2 \\ 
            \frac{\hat{p_i}}{\left | \hat{p_i} \right |}\left \lceil log_{2}\left | \hat{p_i} \right | +1\right \rceil & \left | \hat{p_i} \right |>2  \\ 
    \end{matrix}\right.
\end{equation}
After getting the relative positions $p_i^s$ and $p_i^o$ to the two entities of interest (subject and object respectively), we map them to position embeddings base on a shared position embedding matrix $\mathbf{W}^p$. The two embeddings are concatenated to form the final position embedding for token $x_i$: $\mathbf{p}_i = [\mathbf{p}_i^s, \mathbf{p}_i^o]$.

Position-aware attention is performed on the outputs of knowledge-attention  $\mathbf{O}\in \mathbb{R}^{n\times d_k}$, taking the corresponding relative position embeddings $\mathbf{P}\in \mathbb{R}^{n\times d_p}$ into consideration:
\begin{equation} \label{pos_attn}
    \mathbf{f} = O^Tsoftmax(tanh(\mathbf{O}\mathbf{W}_o+\mathbf{P}\mathbf{W}_p)\mathbf{c})
\end{equation}
where $\mathbf{W}_o\in \mathbb{R}^{d_k\times d_a}$, $\mathbf{W}_p\in \mathbb{R}^{d_p\times d_a}$, $d_a$ is attention dimension, and $\mathbf{c}\in \mathbb{R}^{d_a}$ is a context vector learned by the neural network.

\begin{figure*}
\centering
\includegraphics[width=0.8\linewidth]{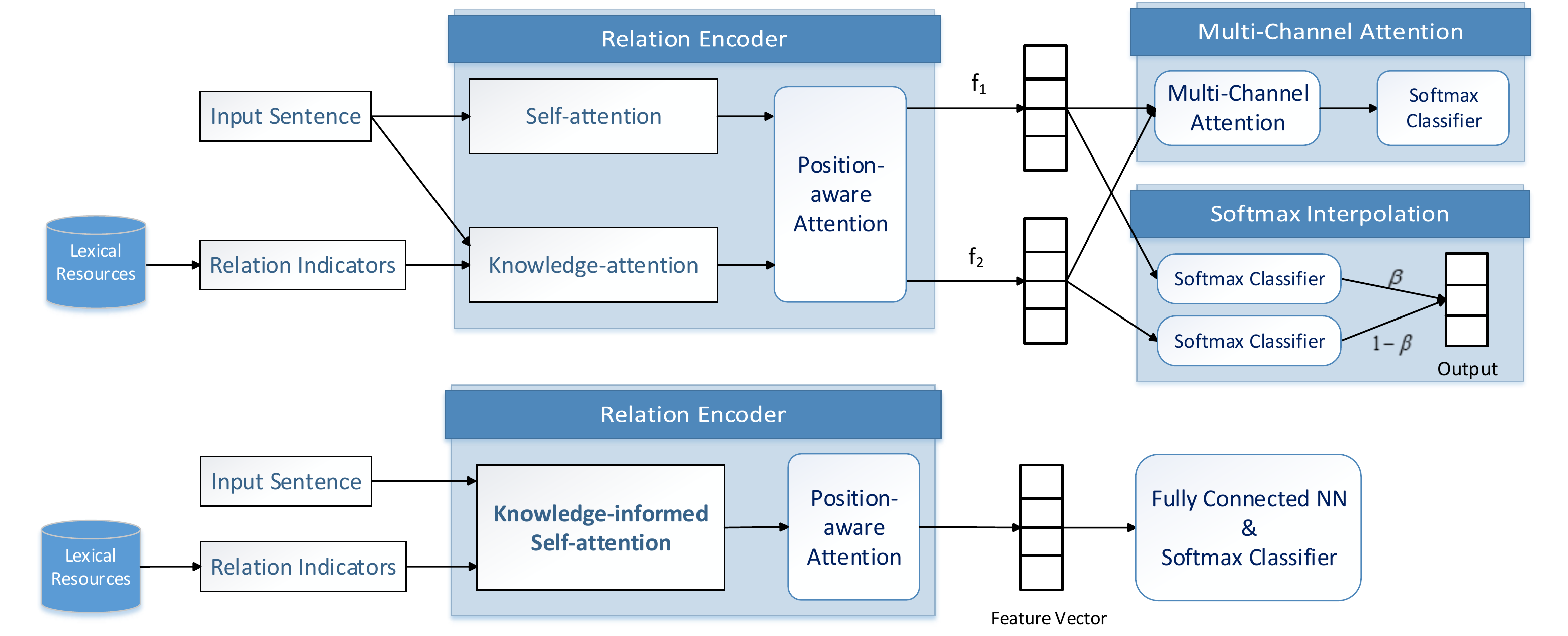}
\caption{Three ways of integrating knowledge-attention with self-attention: multi-channel attention and softmax interpolation (top), as well as knowledge-informed self-attention (bottom).}
\label{fig_combination}
\end{figure*}

\section{Integrate Knowledge-attention with Self-attention}
The self-attention encoder proposed by Vaswani et al.~\shortcite{vaswani2017attention} learns internal semantic features by modeling pair-wise interactions within the texts themselves, which is effective in capturing long-distance dependencies.
Our proposed knowledge-attention encoder has complementary strengths of capturing the linguistic clues of relations precisely based on external knowledge. Therefore, it is beneficial to integrate the two models to maximize the utilization of both external knowledge and training data.
In this section, we propose three integration approaches as shown in Figure \ref{fig_combination}, and each approach has its own advantages.

\subsection{Multi-channel Attention}
In this approach, self-attention and knowledge-attention are treated as two separate channels to model sentence from different perspectives. After applying position-aware attention, two feature vectors $\mathbf{f}_1$ and $\mathbf{f}_2$ are obtained from self-attention and knowledge-attention respectively. We apply another attention mechanism called multi-channel attention to integrate the feature vectors.

In multi-channel attention, feature vectors are first fed into a fully connected neural network to get their hidden representations $\mathbf{h}_i$. Then attention weights are calculated using a learnable context vector $\mathbf{c}$, which reflects the importance of each feature vector to final relation classification. Finally, the feature vectors are integrated based on attention weights, as shown in Equation \ref{combine}.
\begin{equation}\label{combine}
    \mathbf{r} = \sum_i softmax(\mathbf{h}_i^T\mathbf{c}) \mathbf{h}_i
\end{equation}
After obtaining the integrated feature vector $\mathbf{r}$, we pass it to a softmax classifier to determine the relation class. The model is trained using stochastic gradient descent with momentum and learning rate decay to minimize the cross-entropy loss.

The main advantage of this approach is flexibility. Since the two channels process information independently, the input components are not necessary to be the same. Besides, we can add more features from other sources (e.g. subject and object categories) to multi-channel attention to make final decision based on all the information sources.

\subsection{Softmax Interpolation}
Similar as multi-channel attention, we also use two independent channels for self-attention and knowledge-attention in softmax interpolation. Instead of integrating the feature vectors, we make two independent predictions using two softmax classifiers based on the feature vectors from the two channels. 
The loss function is defined as total cross-entropy loss of the two classifiers.
The final prediction is obtained using an interpolation function of the two softmax distributions:
\begin{equation}
    \mathbf{p}=\beta \cdot \mathbf{p}_1 + (1-\beta)\cdot \mathbf{p}_2
\end{equation}
where $\mathbf{p}_1$, $\mathbf{p}_2$ are the softmax distributions obtained form self-attention and knowledge-attention respectively, and $\beta$ is the priority weight assigned to self-attention.

Since knowledge-attention focuses on capturing the keywords and cue phrases of relations, the precision will be higher than self-attention while the recall is lower. The proposed softmax interpolation approach is able to take the advantages of both attention mechanisms and balance the precision and recall by adjusting the priority weight $\beta$. 

\subsection{Knowledge-informed Self-attention}
\begin{figure}
    \centering
    \includegraphics[width=0.8\linewidth]{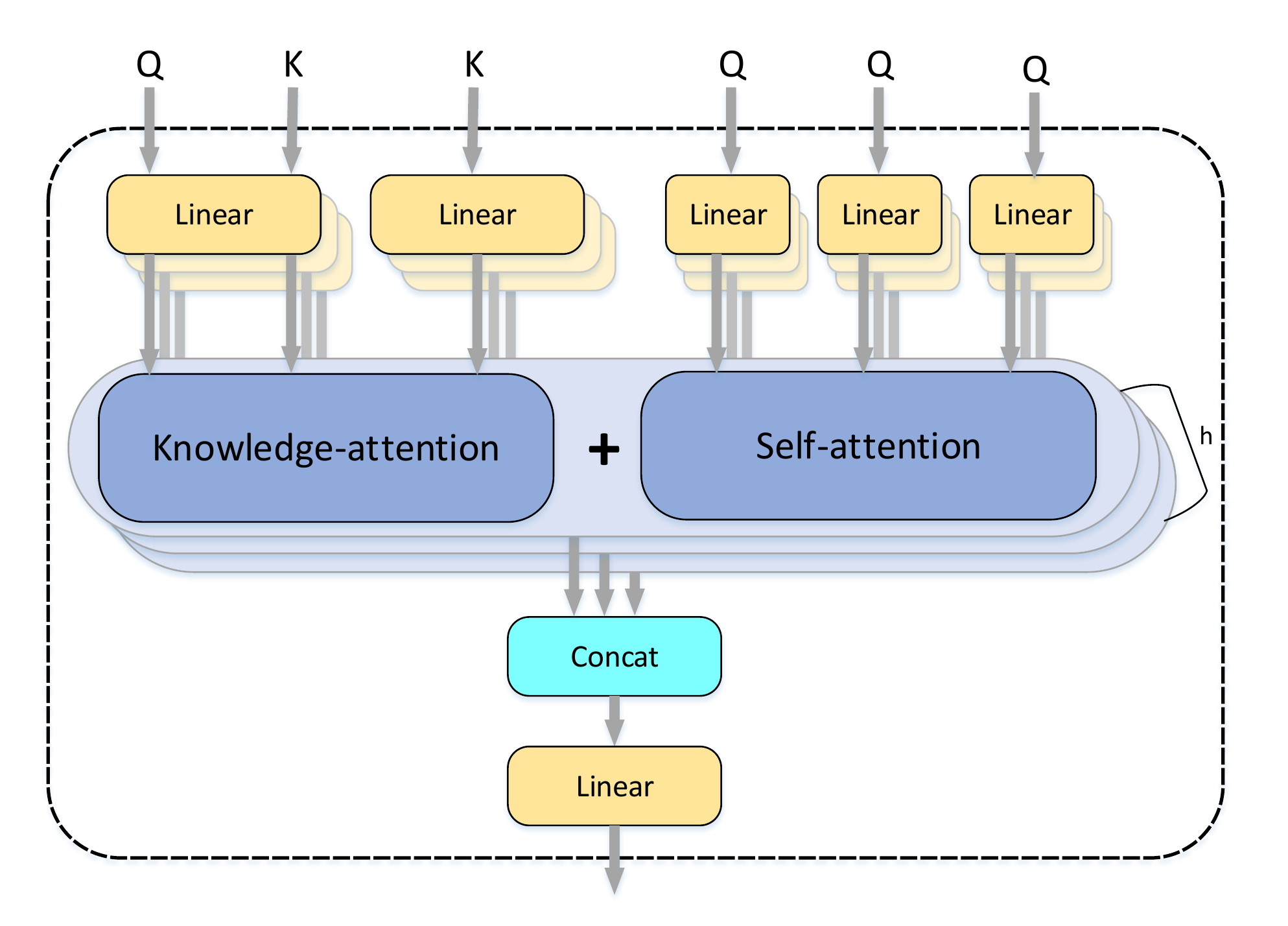}
    \caption{Knowledge-informed self-attention structure. Q, K represent input matrix and knowledge indicators respectively, h is the number of attention heads.}
    \label{fig3}
\end{figure}

Since knowledge-attention and self-attention share similar structures, it is also possible to integrate them into a single channel. We propose knowledge-informed self-attention encoder which incorporates knowledge-attention into every self-attention head to jointly model the semantic relations based on both knowledge and data.

The structure of knowledge-informed self-attention is shown in Figure \ref{fig3}. Formally, given texts input matrix $\mathbf{Q} \in \mathbb{R}^{n\times d_k}$ and knowledge indicators $\mathbf{K}\in \mathbb{R}^{m\times d_k}$. The output of each attention head is calculated as follows:
\begin{equation}
    \begin{matrix}
    head_i=&knwl(\mathbf{Q}\mathbf{W}_i^Q, \mathbf{K}\mathbf{W}_i^Q, \mathbf{K}\mathbf{W}_i^V ) + \\ 
     &self(\mathbf{Q}\mathbf{W}_i^{Q_s}, \mathbf{Q}\mathbf{W}_i^{K_s}, \mathbf{Q}\mathbf{W}_i^{V_s})
    \end{matrix}
\end{equation}
where $knwl$ and $self$ indicate knowledge-attention and self-attention respectively, and all the linear transformation weight matrices have the dimensionality of $\mathbf{W}\in \mathbb{R}^{d_k\times (d_k/h)}$.

Since each self-attention head is aided with prior knowledge in knowledge-attention, the knowledge-informed self-attention encoder is able to capture more lexical and semantic information than single attention encoder.

\section{Experiment and Analysis}
\subsection{Baseline Models}
To study the performance of our proposed models, the following baseline models are used for comparison:

\noindent {\textbf{CNN-based models}} including: (1) CNN: the classical convolutional neural network for sentence classification \cite{kim2014convolutional}. (2) CNN-PE: CNN with position embeddings dedicated for relation classification \cite{nguyen2015relation}. (3) GCN: a graph convolutional network over the pruned dependency trees of the sentence \cite{zhang2018graph}.

\noindent {\textbf{RNN-based models}} including: (1) LSTM: long short-term memory network to sequentially model the texts. Classification is based on the last hidden output.  (2) PA-LSTM: Similar position-aware attention mechanism as our work is used to summarize the LSTM outputs \cite{zhang2017position}. 

\noindent {\textbf{CNN-RNN hybrid model}} including contextualized GCN (C-GCN) where the input vectors are obtained using bi-directional LSTM network~\cite{zhang2018graph}.

\noindent {\textbf{Self-attention-based model}} (Self-attn) which uses self-attention encoder to model the input sentence. Our implementation is based on Bilan and Roth~\shortcite{bilan2018position} where several modifications are made on the original Transformer encoder, including the use of relative positional encodings instead of absolute sinusoidal encodings, as well as other configurations such as residual connection, activation function and normalization. 

For our model, we evaluate both the proposed knowledge-attention encoder (Knwl-attn) as well as the integrated models with self-attention including multi-channel attention (MCA), softmax interpolation (SI) and knowledge-informed self-attention (KISA).

\subsection{Experiment Settings}
We conduct our main experiments on TACRED, a large-scale relation extraction dataset introduced by Zhang et al.~\shortcite{zhang2017position}. TACRED contains over 106k sentences with hand-annotated subject and object entities as well as the relations between them. It is a very complex relation extraction dataset with 41 relation types and a \emph{no\_relation} class when no relation is hold between entities. 
The dataset is suited for real-word relation extraction since it is unbalanced with 79.5\% \emph{no\_relation} samples, and multiple relations between different entity pairs can be exist in one sentence. Besides, the samples are normally long sentences with an average of 36.2 words.

Since the dataset is already partitioned into train (68124 samples), dev (22631 samples) and test (15509 samples) sets, we tune model hyper-parameters using dev set and evaluate model using test set. The evaluation metrics are micro-averaged precision, recall and F$_1$ score. For fair comparison, we select the model with median F$_1$ score on dev set from 5 independent runs, same as Zhang et al.~\shortcite{zhang2017position}.
The same ``entity mask'' strategy is used which replaces subject (or object) entity with special \big \langle NER\big \rangle -SUBJ (or \big \langle NER\big \rangle-OBJ) tokens to avoid overfittting on specific entities and provide entity type information. 

Besides TACRED, another dataset called SemEval2010-Task8~\cite{hendrickx2009semeval} is used to evaluate the generalization ability of our proposed model. The dataset is significantly smaller and simpler than TACRED, which has 8000 training samples and 2717 testing samples. It contains 9 directed relations and 1 other relation (19 relation classes in total). We use the official macro-averaged F$_1$ score as evaluation metric.

We use one layer encoder with 6 attention heads for both knowledge-attention and self-attention since further increasing the number of layers and attention heads will degrade the performance. For softmax interpolation, we choose $\beta=0.8$ to balance precision and recall. 
Word embeddings are fine-tuned based on pre-trained GloVe~\cite{pennington2014glove} with dimensionality of 300. Dropout~\cite{srivastava2014dropout} is used during trianing to alleviate overfitting. Other model hyper-parameters and training details are described in Appendix due to space limitations.

\subsection{Results and Analysis}

\subsubsection{Results on TACRED dataset}
Table \ref{table1} shows the results of baseline as well as our proposed models on TACRED dataset.  It is observed that our proposed knowledge-attention encoder outperforms all CNN-based and RNN-based models by at least 1.3 F$_1$. Meanwhile, it achieves comparable results with C-GCN and self-attention encoder, which are the current start-of-the-art single-model systems.

Comparing with self-attention encoder, it is observed that knowledge-attention encoder results in higher precision but lower recall. This is reasonable since knowledge-attention encoder focuses on capturing the significant linguistic clues of relations based on external knowledge, it will result in high precision for the predicted relations similar to rule-based systems. 
Self-attention encoder is able to capture more long-distance dependency features by learning from data, resulting in better recall. By integrating self-attention and knowledge-attention using the proposed approaches, a more balanced precision and recall can be obtained, suggesting the complementary effects of self-attention and knowledge-attention mechanisms. 
The integrated models improve performance by at least 0.9 F$_1$ score and achieve new state-of-the-art results among all the single end-to-end models.

\begin{table}[t]
\centering
\begin{tabular}{llll}
  \hline 
  Model & P & R & F$_1$\\
  \hline
  CNN$^\dagger$ & \textbf{72.1} & 50.3 & 59.2 \\
  CNN-PE$^\dagger$ & 68.2 & 55.4 & 61.1 \\
  GCN$^\ddagger$ & 69.8 & 59.0 & 64.0\\
  LSTM$^\dagger$ & 61.4 & 61.7 & 61.5 \\
  PA-LSTM$^\dagger$ & 65.7 & 64.5 & 65.1\\
  C-GCN$^\ddagger$ & 69.9 & 63.3 & 66.4\\
  Self-attn$^{\dagger\dagger}$& 64.6 & \textbf{68.6} & 66.5\\
  \hline
  Knwl-attn & 70.0 & 63.1 & 66.4 \\
  Knwl+Self (MCA) & 68.4 & 66.1 & 67.3* \\
  Knwl+Self (SI) & 67.1 & 68.4 & \textbf{67.8}* \\
  Know+Self (KISA) & 69.4 & 66.0 & 67.7* \\
  \hline
\end{tabular}
\caption{Micro-averaged precision (P), recall (R) and F$_1$ score on TACRED dataset. $\dagger$, $\ddagger$ and $\dagger\dagger$ mark the results reported in \cite{zhang2017position}, \cite{zhang2018graph} and \cite{bilan2018position} respectively. $*$ marks statistically significant improvements over Self-attn with $p<0.01$ under one-tailed t-test.}
\label{table1}
\end{table}

\begin{figure}[]
    \centering
    \includegraphics[width=\linewidth]{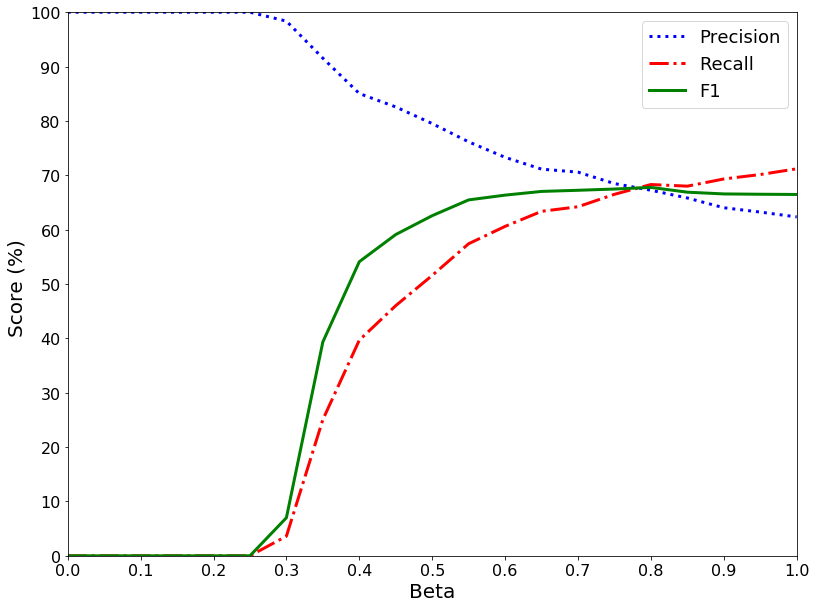}
    \caption{Change of precision, recall and F$_1$ score on dev set as the priority weight $\beta$ in softmax interpolation changes.}
    \label{fig4}
\end{figure}

Comparing the three integrated models, softmax interpolation (SI) achieves the best performance. More interestingly, we found that the precision and recall can be controlled by adjusting the priority weight $\beta$. Figure \ref{fig4} shows impact of $\beta$ on precision, recall and F$_1$ score. As $\beta$ increases, precision decreases and recall increases. Therefore, we can choose a small $\beta$ for relation extraction system which requires high precision, and a large $\beta$ for the system requiring better recall. F$_1$ score reaches the highest value when precision and recall are balanced ($\beta=0.8$).

Knowledge-informed self-attention (KISA) has comparable performance with softmax interpolation, and without the need of hyper-parameter tuning since knowledge-attention and self-attention are integrated into a single channel. The performance gain over self-attention encoder is 1.2 F$_1$ with much improved precision, demonstrating the effectiveness of incorporating knowledge-attention into self-attention to jointly model the sentence based on both knowledge and data.

Performance gain is the lowest for multi-channel attention (MCA). However, the model is more flexible in the way that features from other information sources can be easily added to the model to further improve its performance. Table \ref{table2} shows the results of adding NER embeddings of each token to self-attention channel, and entity (subject and object) categorical embeddings to multi-channel attention as additional feature vectors. We use dimensionality of 30 and 60 for NER and entity categorical embeddings respectively, and the two embedding matrixes are learned by the neural network. 
Results show that adding NER and entity categorical information to MCA integrated model  improves F$_1$ score by 0.2 and 0.5 respectively, and adding both improves precision significantly, resulting a new best F$_1$ score.

\begin{table}[]
\centering
\begin{tabular}{llll}
  \hline 
  Model & P & R & F$_1$\\
  \hline
  MCA & 68.4 & 66.1 & 67.3 \\
  $+$ NER & 68.7 & \textbf{66.2} & 67.5 \\
  $+$ Entity category & 70.0 & 65.8 & 67.8 \\
  $+$ Both & \textbf{70.1} & 66.0 & \textbf{67.9} \\
  \hline
\end{tabular}
\caption{Results of adding NER embeddings and entity categorical embeddings to the multi-channel attention (MCA) integrated model.}
\label{table2}
\end{table}

\subsubsection{Results on SemEval2010-Task8 dataset}
We use SemEval2010-Task8 dataset to evaluate the generalization ability of our proposed model. Experiments are conducted in two manners: mask or keep the entities of interest. Results in Table \ref{table_semveal} show that the ``entity mask'' strategy degrades the performance, indicating that there exist strong correlations between entities of interest and relation classes in SemEval2010-Task8 dataset. Although the results of keeping the entities are better, the model tends to remember these entities instead of focusing on learning the linguistic clues of relations. This will result in bad generalization for sentences with unseen entities.

Regardless of whether the entity mask is used, by incorporating knowledge-attention mechanism, our model improves the performance of self-attention by a statistically significant margin, especially the softmax interpolation integrated model. The results on SemEval2010-Task8 are consistent with that of TACRED, demonstrating the effectiveness and robustness of our proposed method.

\begin{table}[t]
\centering
\begin{tabular}{lcc}
  \hline 
  Model & mask entity & keep entity \\
  \hline
  Self-attn & 76.8\enspace & 83.1\enspace  \\
  Knwl-attn & 76.1\enspace & 82.3\enspace  \\
  Knwl+Self (MCA) & 77.4* & 84.0*  \\
  Knwl+Self (SI) & \textbf{78.0}* & \textbf{84.3}* \\
  Know+Self (KISA) & 77.5* & 84.0* \\
  \hline
\end{tabular}
\caption{Macro-averaged F$_1$ score on SemEval2010-Task8 dataset. $*$ marks statistically significant improvements over Self-attn with $p<0.01$ under one-tailed t-test.}
\label{table_semveal}
\end{table}

\begin{table}[]
\centering
\begin{tabular}{lc}
  \hline 
  Model & Dev F$_1$\\
  \hline
  Knwl-attn Encoder & 66.5  \\
  1. $-$ Multi-head structure & 64.6  \\
  2. $-$ Synonym relation indicators & 64.7  \\
  3. $-$ Relation indicators mean & 65.0  \\
  4. $-$ Output masking & 65.8 \\
  5. $-$ Entity masking & 65.4  \\
  6. $-$ Relative positions & 63.0  \\
  \hline
\end{tabular}
\caption{Ablation study on knowledge-attention encoder. Results are the median F$_1$ scores of 5 independent runs on dev set of TACRED.}
\label{table3}
\end{table}

\subsubsection{Ablation study}

To study the contributions of specific components of knowledge-attention encoder, we perform ablation experiments on the dev set of TACRED. The results of knowledge-attention encoder with and without certain components are shown in Table \ref{table3}. 

It is observed that: (1) The proposed multi-head knowledge-attention structure outperforms single-head significantly. This demonstrates the effectiveness of jointly attending texts to different relational semantic subspaces in the multi-head structure. (2) The synonyms improve the performance of knowledge-attention since they are able to broaden the coverage of relation indicators and form a robust relational semantic space. (3) The subtraction of relation indicators mean vector from attention hidden representations helps to suppress the activation of
irrelevant words and results in a better representation for each word to capture the linguistic clues of relations. (4-5) The two masking strategies are helpful for our model: the output masking eliminates the effects of the padding tokens and the entity masking avoids entity overfitting while providing entity type information. (6) The relative position embedding term in position-aware attention contributes a significant amount of F$_1$ score. This shows that positional information is particularly important for relation extraction task.

\begin{table*}[t]
  \centering
  \begin{tabular}{  m{11.9cm}  p{2cm}  m{0.8cm}  }
    \hline
    \small Sample Sentences & \small True Relation & \small Predict \\ \hline
    
    \begin{minipage}{0.75\textwidth}
      \includegraphics[width=\linewidth, trim=0 0 0 -1]{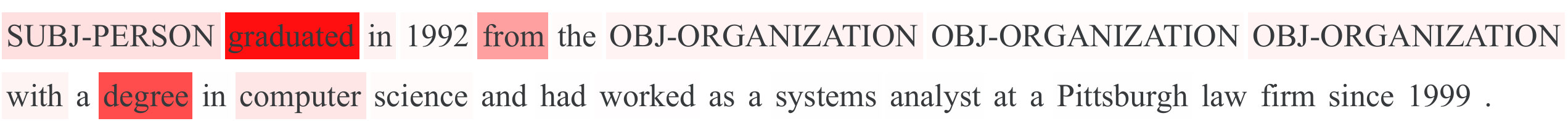}
    \end{minipage}
    & \small per:schools \textunderscore attended
    & \small correct \\ 
    \begin{minipage}{0.75\textwidth}
      \includegraphics[width=\linewidth]{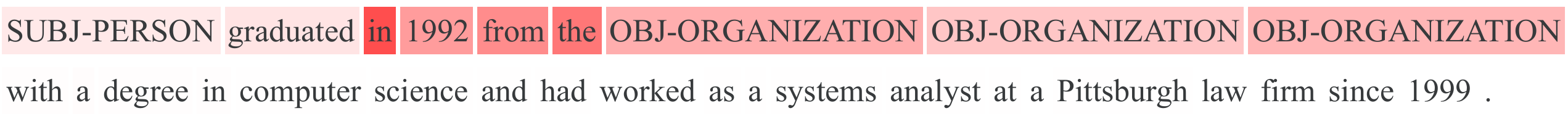}
    \end{minipage}
    & 
    & \small correct
    \\ \hline
    
    \begin{minipage}{0.75\textwidth}
      \includegraphics[width=\linewidth, trim=0 0 0 -1]{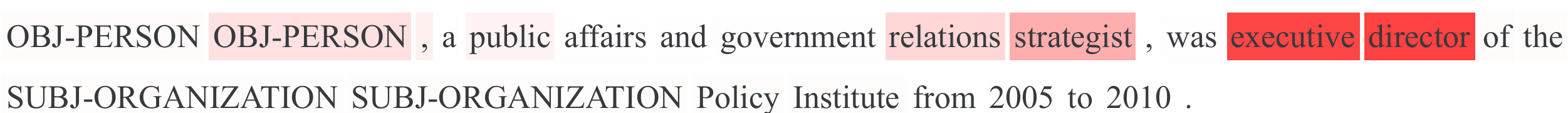}
    \end{minipage}
    & \small org:top\textunderscore members /employees
    & \small correct \\ 
    \begin{minipage}{0.75\textwidth}
      \includegraphics[width=\linewidth]{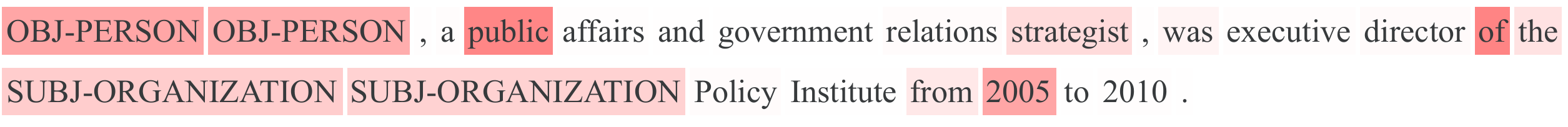}
    \end{minipage}
    & 
    & \small wrong
    \\ \hline

    \begin{minipage}{0.75\textwidth}
      \includegraphics[width=\linewidth, trim=0 0 0 -1]{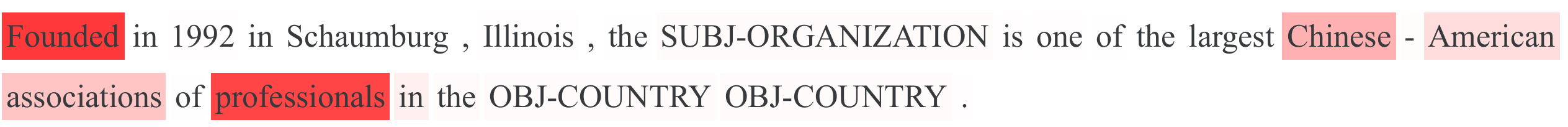}
    \end{minipage}
    & \small org:country\textunderscore of \textunderscore headquarters
    & \small wrong \\ 
    \begin{minipage}{0.75\textwidth}
      \includegraphics[width=\linewidth]{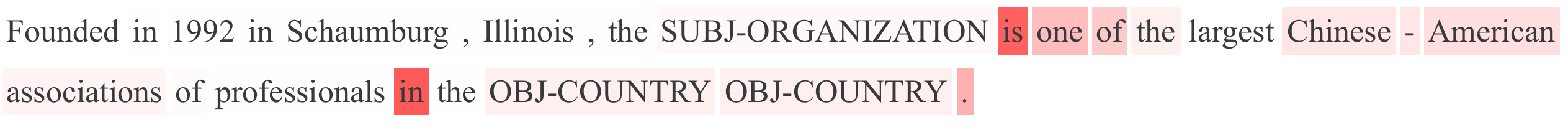}
    \end{minipage}
    & 
    & \small correct
    \\ \hline

  \end{tabular}
  \caption{Attention visualization for knowledge-attention encoder (first) and self-attention encoder (second). Words are highlighted based on the attention weights assigned to them. Best viewed in color.}
  \label{table4}
\end{table*}

\subsubsection{Attention visualization}
To verify the complementary effects of knowledge-attention encoder and self-attention encoder, we compare the attention weights assigned to words from the two encoders. Table \ref{table4} presents the attention visualization results on sample sentences. For each sample sentence, attention weights from knowledge-attention encoder are visualized first, followed by self-attention encoder. 
It is observed that knowledge-attention encoder focuses more on the specific keywords or cue phrases of certain relations, such as ``graduated", ``executive director" and ``founded";
while self-attention encoder attends to a wide range of words in the sentence and pays more attention to the surrounding words of target entities especially the words indicating the syntactic structure, such as ``is", ``in" and ``of". 
Therefore, knowledge-attention encoder and self-attention encoder have complementary strengths that focus on different perspectives for relation extraction.

\subsubsection{Error analysis}
To investigate the limitations of our proposed model and provide insights for future research, we analyze the errors produced by the system on the test set of TACRED. 
For knowledge-attention encoder, 58\% errors are false negative (FN) due to the limited ability in capturing long-distance dependencies and some unseen linguistic clues during training.
For our integrated model\footnote{We observed similar error behaviors of the three proposed integrated models.} that takes the benefits of both self-attention and knowledge-attention, FN is reduced by 10\%. However, false positive (FP) is not improved due to overfitting that leads to wrong predictions. Many errors are caused by multiple entities with different relations co-occurred in one sentence. Our model may mistake irrelevant entities as a relation pair.
We also observed that many FP errors are due to the confusions between related relations such as ``city\_of\_death''and ``city\_of\_residence". More data or knowledge is needed to distinguish ``death'' and ``residence''. Besides, some errors are caused by imperfect annotations.

\section{Conclusion and Future Work}
We introduce knowledge-attention encoder which effectively incorporates prior knowledge from external lexical resources for relation extraction. The proposed knowledge-attention mechanism transforms texts from word space into relational semantic space and captures the informative linguistic clues of relations effectively. Furthermore, we show the complementary strengths of knowledge-attention and self-attention, and propose three different ways of integrating them to maximize the utilization of both knowledge and data. The proposed models are fully attention-based end-to-end systems and achieve state-of-the-art results on TACRED dataset, outperforming existing CNN, RNN, and self-attention based models.

In future work, besides lexical knowledge, we will incorporate conceptual knowledge from encyclopedic knowledge bases into knowledge-attention encoder to capture the high-level semantics of texts. We will also apply knowledge-attention in other tasks such as text classification, sentiment analysis and question answering.



\bibliography{emnlp2019}
\bibliographystyle{acl_natbib}

\onecolumn
\appendix

\section{FramNet Frames Used for Relation Extraction on TACRED Dataset}
\label{sec_supplementalA}

\begin{table}[h]
\centering

\begin{tabular}{AB}
  \hline 
  Relation Types & FrameNet Frames\\
  \hline
  org:alternate\textunderscore names, per:alternate\textunderscore names 
  & Being\textunderscore named, Name\textunderscore conferral, Namesake, Referring\textunderscore by\textunderscore name, Simple\textunderscore naming  \\
  \hline
  org:city\textunderscore of\textunderscore headquarters, org:country\textunderscore of\textunderscore headquarters, org:stateorprovince\textunderscore of\textunderscore headquarters 
  & Being\textunderscore located, Locale, Locale\textunderscore by\textunderscore characteristic\textunderscore entity, Locale\textunderscore by\textunderscore collocation, Locale\textunderscore by\textunderscore event, Locale\textunderscore by\textunderscore ownership, Locale\textunderscore by\textunderscore use, Locale\textunderscore closure, Locating, Locative\textunderscore relation, Spatial\textunderscore co-location  \\
  \hline
  org:founded, org:founded\textunderscore by
  & Intentionally\textunderscore create \\
  \hline
  org:dissolved
  & Location\textunderscore in\textunderscore time, Relative\textunderscore time, Time\textunderscore vector, Timespan, Temporal\textunderscore collocation, Temporal\textunderscore subregion \\
  \hline 
  org:member\textunderscore of,  org:members
  & Becoming\textunderscore a\textunderscore member, Membership \\
  \hline 
  org:political/religious\textunderscore affiliation, per:religion
  & People\textunderscore by\textunderscore religion, Religious\textunderscore belief, Political\textunderscore locales \\
  \hline
  org:subsidiaries, org:parents
  & Part\textunderscore whole, Partitive, Inclusion \\
  \hline 
  org:shareholders
  & Capital\textunderscore stock \\
  \hline
  org:top\textunderscore members/employees, org:number\textunderscore of\textunderscore employees/members, per:employee\textunderscore of 
  & Leadership, Working\textunderscore a\textunderscore post, Employing, People\textunderscore by\textunderscore vocation, Cardinal\textunderscore numbers \\
  \hline 
  per:age
  & Age \\
  \hline 
  per:cause\textunderscore of\textunderscore death, per:date\textunderscore of\textunderscore death, per:city\textunderscore of\textunderscore death, per:country\textunderscore of\textunderscore death, per:stateorprovince\textunderscore of\textunderscore death
  & Death, Cause\textunderscore harm \\
  \hline 
  per:charges
  & Notification\textunderscore of\textunderscore charges, Committing\textunderscore crime, Criminal\textunderscore investigation \\
  \hline 
  per:children, per:parents, per:other\textunderscore family, per:siblings
  & Kinship \\
  \hline
  per:cities\textunderscore of\textunderscore residence, per:countries\textunderscore of\textunderscore residence, per:stateorprovinces\textunderscore of\textunderscore residence
  & Expected\textunderscore location\textunderscore of\textunderscore person, Residence \\
  \hline
  per:date\textunderscore of\textunderscore birth, per:city\textunderscore of\textunderscore birth, per:country\textunderscore of\textunderscore birth, per:stateorprovince\textunderscore of\textunderscore birth
  & Being\textunderscore born, Giving\textunderscore birth \\
  \hline 
  per:origin 
  & Origin, People\textunderscore by\textunderscore origin \\
  \hline 
  per:schools\textunderscore attended
  & Education\textunderscore teaching \\
  \hline 
  per:spouse 
  & Personal\textunderscore relationship, Forming\textunderscore relationships \\
  \hline 
  per:title 
  & Performers\textunderscore and\textunderscore roles \\
  \hline
\end{tabular}
\label{appendix_a}
\end{table}

\section{Hyper-parameter Settings and Training Details}
\label{sec_supplementalB}
The dimension of POS and position embeddings are both 30. The inner layer dimension in position-wise feed-forward network is 130. The dimension of the relative positional encoding within knowledge-attention and self-attention is 50. The attention dimension is 200 in position-aware attention and 100 in multi-channel attention. The fully connected network before softmax has a dimensionality of 100.  We use ReLU for all the nonlinear activation functions.
Dropout rate is 0.4 for input embeddings, attention outputs and position-wise feed-forward outputs, and 0.1 for attention weights dropout. 

The models are trained using Stochastic Gradient Descent with learning rate of 0.1 and momentum of 0.9. The learning rate is decayed with a rate of 0.9 after 15 epochs if F$_1$ score on dev set does not improve.
The batch size is set to 100 and we train the model for 70 epochs.

\end{document}